\documentclass[10pt,twocolumn,letterpaper]{article}

\usepackage[pagenumbers]{cvpr} % To force page numbers, e.g. for an arXiv version

\usepackage{marvosym}
\usepackage{graphicx}
\usepackage{amsmath}
\usepackage{amssymb}
\usepackage{amsfonts}
\usepackage{booktabs}
\usepackage{stfloats}
\usepackage{float}
\usepackage{stfloats}
\def\ie{\emph{i.e}\onedot}
\usepackage[pagebackref,breaklinks,colorlinks]{hyperref}

\usepackage[capitalize]{cleveref}
\crefname{section}{Sec.}{Secs.}
\Crefname{section}{Section}{Sections}
\Crefname{table}{Table}{Tables}
\crefname{table}{Tab.}{Tabs.}

\def\cvprPaperID{6139} % *** Enter the CVPR Paper ID here
\def\confName{CVPR}
\def\confYear{2022}

\begin{document}

%%%%%%%%% TITLE - PLEASE UPDATE
\title{Self-supervised Correlation Mining Network for Person Image Generation }

\author{Zijian Wang\textsuperscript{1}, Xingqun Qi\textsuperscript{1}, Kun Yuan\textsuperscript{2}, Muyi Sun \textsuperscript{3(\Letter)}   \\
\textsuperscript{1}School of Automation, Beijing University of Posts and Telecommunications\\
\textsuperscript{2}Kuaishou Technology
\textsuperscript{3}CRIPAC, Institute of Automation, Chinese Academy of Sciences\\
{\tt\small \{wangzijianbupt, XingqunQi\}@bupt.edu.cn, yuankun03@kuaishou.com, muyi.sun@cripac.ia.ac.cn }
}

% \author{Zijian Wang, Xingqun Qi\\
% School of Automation, BUPT\\
% {\tt\small wangzijianbupt@bupt.edu.cn,XingqunQi@bupt.edu.cn}
% % For a paper whose authors are all at the same institution,
% % omit the following lines up until the closing ``}''.
% % Additional authors and addresses can be added with ``\and'',
% % just like the second author.
% % To save space, use either the email address or home page, not both
% \and
% Kun Yuan \\
% Kuaishou Technology\\
% {\tt\small  yuankunbupt@gmail.com}
% \and
% MuYi Sun \\
% CRIPAC, Institute of Automation, Chinese Academy of Sciences\\
% {\tt\small  muyi.sun@cripac.ia.ac.cn}
% }

% \begin{figure*}[ht]
%     \centering
%     \includegraphics[scale=0.5]{latex/figure/111111.png}
%     \caption{Caption}
%     \label{fig:my_label}
% \end{figure*}

\twocolumn[{%
\renewcommand\twocolumn[1][]{#1}%
\maketitle
\begin{center}
    \centering
    \includegraphics[scale=0.52]{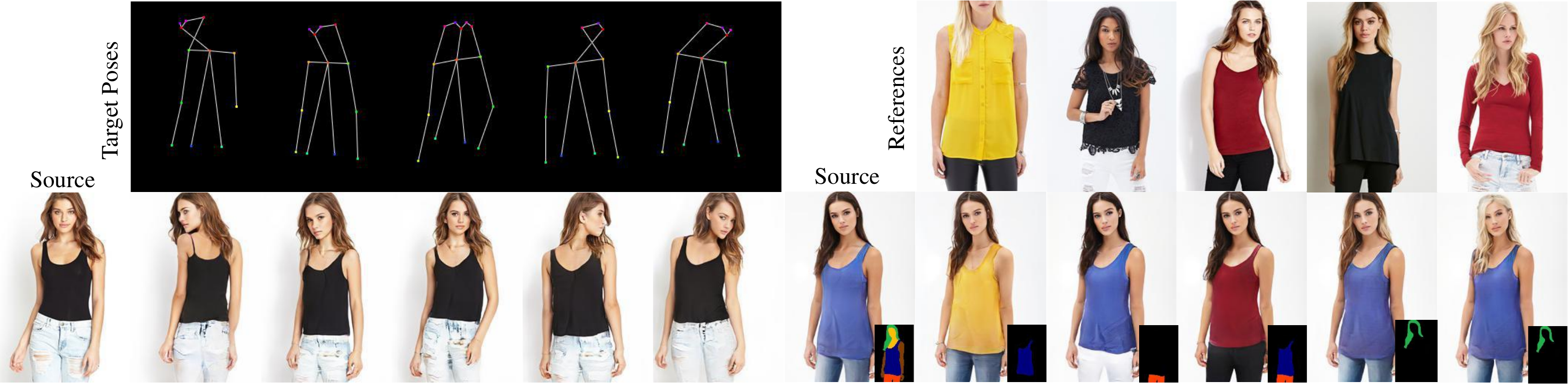}
    \captionof{figure}{
    Exemplary samples synthesized by our self-supervised person image generation framework. 
    Our method could generate the same person images with the target poses (left image set), or generate person images with specific attributes referring to different person images (right image set).  
    Better viewed by zooming in the electronic version.}
    \label{figure1}
\end{center}
}]

%%%%%%%%% ABSTRACT
\begin{abstract}
% Person image generation has shown great potential in many applications, such as film industry and multimedia creation.
% Recently, self-supervised methods express great prospects in this task, which disentangle the style features and merge them with pose features for self-reconstruction. 
% However, such methods merge these features from the global perspective, failing to capture the spatial correlations for person image deformation.
% In this paper, we propose a Self-supervised Correlation Mining Network (SCM-Net) to enhance the spatial correlations in the self-supervised framework. Specifically, we employ a Decomposed Style Encoder (DSE) to construct semantic-decoupled style features. 
% %Then the dense spatial correlations between the style and pose features are established through Correlation Mining Module (CMM).  %sun
% Then we design a Correlation Mining Module to establish the dense spatial correlations between the style and pose features.
% Based on these correlations, the style information could be warped into the pose representations for self-reconstruction. 
% Meanwhile, to improve the fidelity of cross-scale pose transformation, we propose a Body Structure Retaining Loss (BSR Loss) to constrain the person image structure. 
% Extensive experiments conducted on DeepFashion dataset achieve 12.18 on FID, 0.182 on LPIPS, and 3.632 on IS, which demonstrate the superiority of our method compared with other supervised and unsupervised approaches. 
% Furthermore, satisfactory results on face generation tasks show the generalization of our method.
%\vfill

Person image generation aims to perform non-rigid deformation on source images, which generally requires unaligned data pairs for training. 
Recently, self-supervised methods express great prospects in this task by merging the disentangled representations for self-reconstruction. 
However, such methods fail to exploit the spatial correlation between the disentangled features. 
In this paper, we propose a Self-supervised Correlation Mining Network (SCM-Net) to rearrange the source images in the feature space, in which two collaborative modules are integrated, Decomposed Style Encoder (DSE) and Correlation Mining Module (CMM).
Specifically, the DSE first creates unaligned pairs at the feature level. 
Then, the CMM establishes the spatial correlation field for feature rearrangement. 
Eventually, a translation module transforms the rearranged features to realistic results.
Meanwhile, for improving the fidelity of cross-scale pose transformation, we propose a graph based Body Structure Retaining Loss (BSR Loss) to preserve reasonable body structures on half body to full body generation.
Extensive experiments conducted on DeepFashion dataset demonstrate the superiority of our method compared with other supervised and unsupervised approaches. 
Furthermore, satisfactory results on face generation show the versatility of our method in other deformation tasks.
\end{abstract}

%%%%%%%%% BODY TEXT
\section{Introduction}
\label{sec:intro}
Pose guided person image generation is an unaligned image to image translation problem, which aims to change the posture of a person image given target poses as condition \cite{ma2017pose,zhu2019progressive,men2020controllable,tang2020xinggan,zhang2021pise,ma2021must}. 
Person image generation has shown great potential in many fields, such as film industry and multimedia creation.
However, the difficulty of performing non-rigid deformation
makes this task an active topic in the community of computer vision.

Due to the large spatial misalignment between the source and target images, existing approaches generally need unaligned data pairs to supervise the training process\cite{ma2017pose,zhu2019progressive,men2020controllable,tang2020xinggan,zhang2021pise}. 
For instance, \cite{zhu2019progressive,tang2020xinggan} calculate the attention map between paired poses to guide the anomalous pose deformation. \cite{ren2020deep,dong2018soft,han2019clothflow} establish the coordinate offset flow to promote the position-level source feature sampling for person feature alignment.
With such attention or flow mechanism, the generative methods could be capable to perform spatial transformations when the source images and target poses are provided.
However, collecting paired data requires heavy workload and limits the application scenarios of these supervised approaches.
Therefore, some unsupervised methods are proposed to deal with this limitation\cite{pumarola2018unsupervised,song2019unsupervised}, which utilize cycle consistent methods or create pseudo labels to promote the training procedure. 
However, such methods still have limitations in generation quality intuitively.

Recently, self-supervised methods demonstrate powerful prospect to perform non-rigid spatial transformations with only source images\cite{ma2018disentangled, esser2018variational, ma2021must}.
They could learn disentangled representations of different image types, which are merged in the following for self-reconstruction. 
Early studies \cite{ma2018disentangled,esser2018variational} employ multi-branch network to disentangle different features and concatenate them to reconstruct the source images.
Ma \etal\cite{ma2021must} utilize AdaIN\cite{huang2017arbitrary} for feature merging by transferring statistics from style features to pose features. 
However, these methods still encounter three challenges. 
First, the disentangled features are aligned in the feature space, which cannot provide enough supervision for spatial transformation in self-supervised methods. 
Second, these merging methods (\eg concatenation or statistics transfer) are global operations, which are limited to exploit the spatial correlation information. 
Third, the model lacks prior knowledge of invisible regions due to the self-supervised training process within the single pose scale, which limits the reasonable completions for invisible regions in the half body to full body transformation.  

In this paper, we propose a Self-supervised Correlation Mining Network (SCM-Net) for person image generation. 
The entire architecture of SCM-Net can be summarized as disentanglement, fusion and translation. 
In the disentanglement phase, inspired by the decomposed strategy in \cite{men2020controllable}, we design a Decomposed Style Encoder (DSE) to extract the semantic-aware decoupled style features, which could form ``unaligned pairs " with its counterpart pose features. 
Through this design, the source image itself could provide supervision for spatial feature deformation. 
In the fusion phase, we propose a Correlation Mining Module (CMM) to further exploit the spatial correlation between disentangled feature pairs. 
The CMM module computes the pairwise correlation between the corresponding positions of feature pairs to establish the dense spatial correlation field.
% Based on this correlation field, all positions of the style feature map are rearranged spatially. 
% Through this way, the ``unaligned pairs " are aligned in the feature space. 
Based on this correlation field, our model could align these disentangled features through spatially rearranging the style feature positions.
%of the style feature map spatially.
In the translation phase, a translation generator with skip connections is introduced to transform the rearranged style features to realistic person images.
The entire model is trained in an end-to-end manner.

For the lack of prior information on the lower body, we design a Body Structure Retaining Loss (BSR Loss) to capture the semantic relationships among different body parts. 
Thus, the model could make reasonable completions based on these relationships.
Specifically, we employ the graph representation to model the semantic relationships of human body parts. 
In this body graph, each node represents the perceptual features of each semantic region and each edge measures the similarity between each node pair.
We match the graphs between each input person image and the corresponding generated result to establish the graph based constraint, which incorporates the body semantic relationships into our model.

% Specifically, we first separate each semantic attribute from the source images based on the corresponding semantic layouts, which are extracted through a pre-trained human parser\cite{gong2017look}. 
% Then, we construct the specific Body Graph to represent the body structure in which the perceptual features of each body semantic attribute are treated as the graph nodes and the similarity between each node pair as the graph edge. 
% Finally, the constraint called Body Structure Retaining Loss is established between the input person image and generated target results. 
% Under the constraint of this loss, the model can capture more context information for generating reasonable results when dealing with large pose transformation.

% In the training framework of our method, we inherit the self-supervised training strategy\cite{ma2018disentangled,ma2021must} through the self-reconstruction of source images. 
% As a result, our model can introduce new target poses for human pose transfer, and edit attributes in the person images during inference time. The main contributions can be summarized as follows:

During inference, our model could introduce new target poses for human pose transfer, and perform reference based attribute editing through partial replacement of style features. 
Figure \ref{figure1} shows some applications of our model.

The main contributions can be summarized as follows:
% \begin{itemize}

% $\bullet$  We propose a Self-supervised Correlation Mining Network (SCM-Net) to enhance the spatial correlations into the self-supervised person image generation framework, which is a concise but effective method. 
$\bullet$  We propose a Self-supervised Correlation Mining Network (SCM-Net) to achieve person image deformation without the supervision of unaligned data pairs.
% \end{itemize}
% \begin{itemize}

$\bullet$ We design two main collaborative modules, the Decomposed Style Encoder (DSE) and the Correlation Mining Module (CMM), which could perform feature disentangling and merging for person image deformation. 
% \end{itemize}
% \begin{itemize}

$\bullet$ We propose a Body Structure Retaining Loss (BSR Loss) to acquire the prior knowledge of invisible regions through incorporating semantic relationships among body parts.
% \end{itemize}
% \begin{itemize}

$\bullet$ Our method performs competitive results compared with the state-of-the-art methods and also obtains satisfactory results on face generation tasks, which demonstrates the migration capability of our model. 
% \end{itemize}
%-------------------------------------------------------------------------
\section{Related Work}
\subsection{Person Image Generation}
 With the dramatic development of Generative Adversarial Networks(GANs)\cite{goodfellow2014generative}, person image generation have made great progress in recent years\cite{ma2017pose,esser2018variational,zhang2020cross,ma2018disentangled,ma2021must,men2020controllable,pumarola2018unsupervised,ren2020deep,song2019unsupervised,tang2020xinggan,zhu2019progressive,zhang2021pise}. 
 Ma \etal\cite{ma2017pose} first introduced the pose-guide person image generation task and proposed a two stage generator to generate target person image. 
 Zhu \etal\cite{zhu2019progressive,tang2020xinggan} proposed an attention mechanism to transfer the image information from source pose to target pose. 
 Ren \etal\cite{ren2020deep} predicted the flow field between source person images and target poses for generating new pose images. 
 Men \etal\cite{men2020controllable} used decomposed component encoding strategy to achieve pose transfer and person attribute editing. 
 However, all the above methods need paired data to supervise the training process, which would take heavy workload for data collection. 
 Several unsupervised methods have been proposed for person image generation.
 Pumarola \etal\cite{pumarola2018unsupervised} designed a bidirectional generator and employed cycle-consistent method to supervise the training. 
 Song \etal\cite{song2019unsupervised} designed a novel schema to generate pseudo semantic maps for the unsupervised generation. However, these methods still need extra target poses as input and have some artifacts in generated images. 
 Recently, \cite{ma2018disentangled,esser2018variational,ma2021must} proposed self-driven methods to settle these problems. 
 However, these methods have limitations when dealing with large pose deformation problems. 
 Inspired by the above, in this paper, we propose a novel self-supervised framework with graph representation learning for person image generation.

\begin{figure*}[ht]
    \centering
    \includegraphics[scale=0.63]{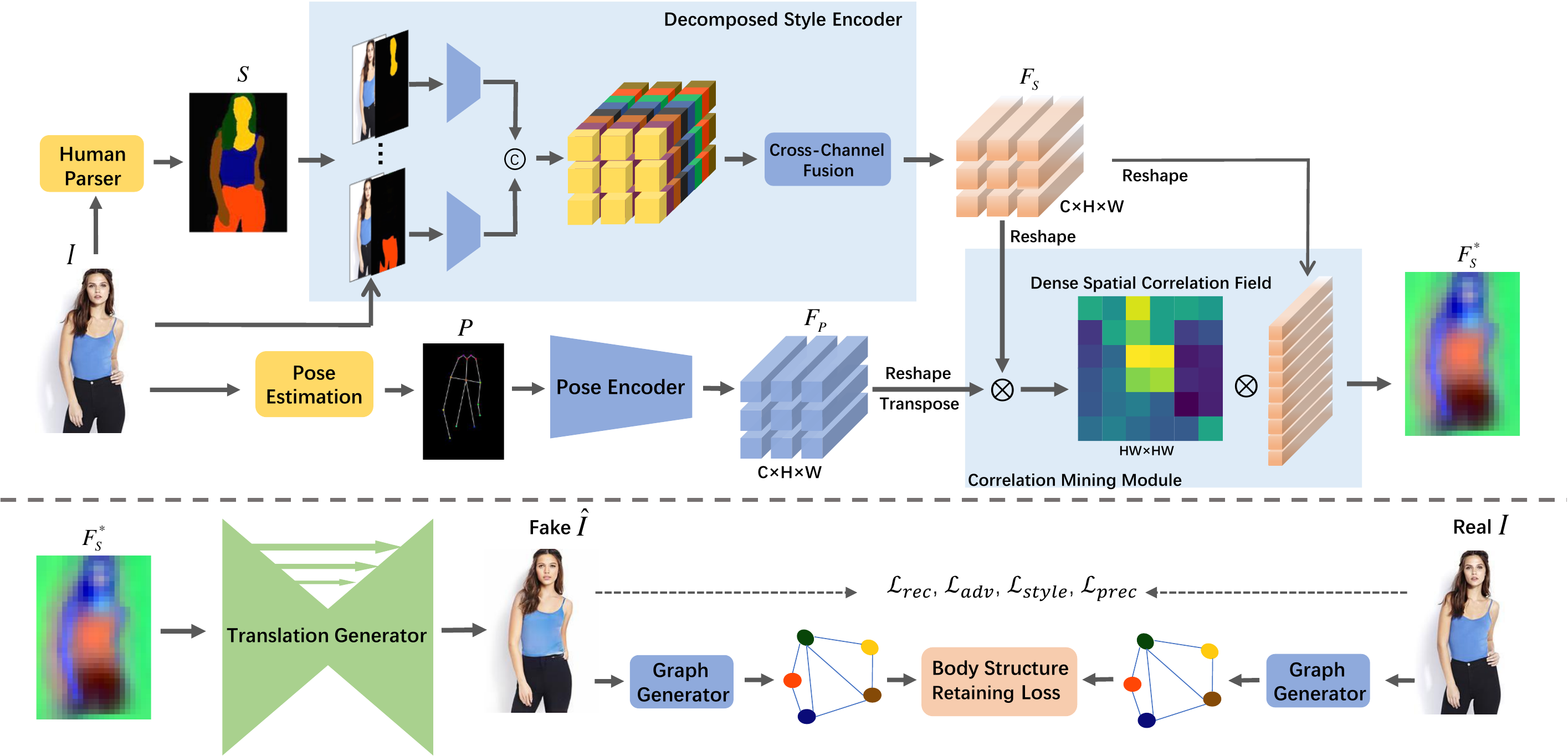}
    \caption{An overall workflow of our self-supervised person image generation framework. Given an input person image $I$, we first utilize the pre-trained methods to obtain its parsing map $S$ and pose skeleton $P$. Then, the Decomposed Style Encoder disentangles the semantic-aware decoupled style features $F_s$ from its pose features $F_p$. Next, the Correlation Mining Module establishes the correlation field $\mathcal{C}$ to guide the feature merging. Finally, the merged features $F_s^*$ are fed into the translation generator to get the reconstruction $\hat{I}$ of the source image. The Body Structure Retaining Loss and other losses are designed to promote the training process. }
    \label{figure2}
\end{figure*}

\subsection{Spatial Correlation Learning}
The purpose of spatial correlation learning is to establish dense spatial correlation fields for image translation. 
% Wang \etal\cite{wang2018non} proposed a non-local operation to capture long-range dependencies in convolutional neural networks for image recognition.
Liao \etal\cite{liao2017visual} proposed a coarse-to-fine strategy to compute the spatial correlation field for image analogy and style transfer. 
He \etal\cite{he2018deep} measured the spatial similarity between the reference and the target to perform exemplar-based colorization. 
Lee \etal\cite{lee2020reference} designed a spatially correlation related module to introduce information from reference image to sketch image for sketch colorization. 
Zhang \etal\cite{zhang2021pise} proposed a spatial-aware normalization module to preserve the spatial context relationship for human pose transfer. 
Zhang \etal\cite{zhang2020cross} established the spatial correlation field in a shared domain to perform cross-domain image to image translation.
% However, these methods are mostly employed for feature representation or cross domain feature alignment.
% %%%%这个地方需要重新写缺点是什么到底，我写的实在是牵强
% In this paper, inspired by \cite{zhang2020cross,zhang2021pise}, we introduce the spatial correlation learning into our self-supervised network to make up for the deficiency of previous feature fusion methods.
However, the above methods can only handle the spatial correlation between unaligned data pairs. In this paper, we establish the correlation field  between the disentangled features of source images, which explores more scenario for spatial correlation learning. 

% Early works on spatial correlation focused on matching hand-crafted features, such as SIFT and HOG\cite{hur2015generalized,liu2010sift,taniai2016joint}. In recent years, deep features extracted by Convolutional Neural Network(CNN) have shown powerful ability for representing semantic information\cite{zeiler2014visualizing,simonyan2014very,he2016deep}. \cite{long2014convnets} first employed CNN features to establish semantic correspondences between images. \cite{he2018deep,zhang2019deep} established the semantic correspondences by calculating the pixel-level similarity between reference images and target gray images to achieve image colorization. Several methods formulated semantic correspondence as a geometric alignment problem to learn the transformation between images\cite{seo2018attentive,rocco2018end,han2017scnet}. \cite{zhang2020cross} aligned the images from different domains to achieve cross domain image to image translation. Our method is similar to \cite{zhang2020cross}, but the main difference is that our method doesn't need paired data to supervise the training process.

\subsection{Graph Representation Learning}
Graph representation learning plays a significant role in the computer vision\cite{yan2018spatial,cao2016deep,zhang2021keypoint}. 
Due to the powerful capabilities of relationship modeling, the graph representation learning has been applied to many tasks, such as skeleton-based action recognition\cite{yan2018spatial}, biometrics recognition\cite{ren2020dynamic} and person re-identification\cite{wu2020adaptive,yan2019learning,shen2018person}.
Yan \etal\cite{yan2019learning} built a person-feature based graph to model the relations among images for person search.
Ren \etal\cite{ren2020dynamic} proposed a dynamic graph for occlusion biometrics recognition. 
Wu \etal\cite{wu2020adaptive} proposed an adaptive graph representation learning scheme to promote the interactions between relevant regional features for video person Re-ID.
Hou \etal\cite{hou2020inter} proposed a graph matching strategy to distill structural knowledge for road marking segmentation. 
Qi \etal\cite{qi2021face} proposed an adaptive re-weighting graph to balance the contributions of different semantic nodes in face sketch synthesis.
% However, these methods are mostly applied on the image representation task with the same image scale, ignoring of the characteristics of graph for cross-scale image complication.
% In this paper, we employ the graph representation into our framework to model the relation among body parts, aiming to generate more reasonable body structures, especially for the half body to full body image generation.
However, the above methods employ graph representation learning to enhance the ability for feature extraction or feature matching, ignoring of the characteristics of graph for cross-scale image complication. 
In this paper, we apply graph representation to model the semantic relationships for cross-scale person image generation, aiming to generate more reasonable body structures.

\section{Method}
In this section, we present our proposed method in detail.
To begin with, we introduce the overall workflow of our Self-supervised Correlation Mining Network (SCM-Net). 
Then we describe the whole network architecture in detail according to the three phases of disentangling, merging and translation. 
Finally, the total objective functions of our model are introduced.
% and introduce the two main collaborative modules integrated in our networks, named as Decomposed Style Encoder(DSE) and Correlation Mining Module (CMM). 
% Afterwards, we present the graph based Body Graph Retaining Loss for body structure preservation. 
% Finally, the total objective functions of our model are introduced.

\begin{figure}[t]
    \centering
    \includegraphics[scale=0.5]{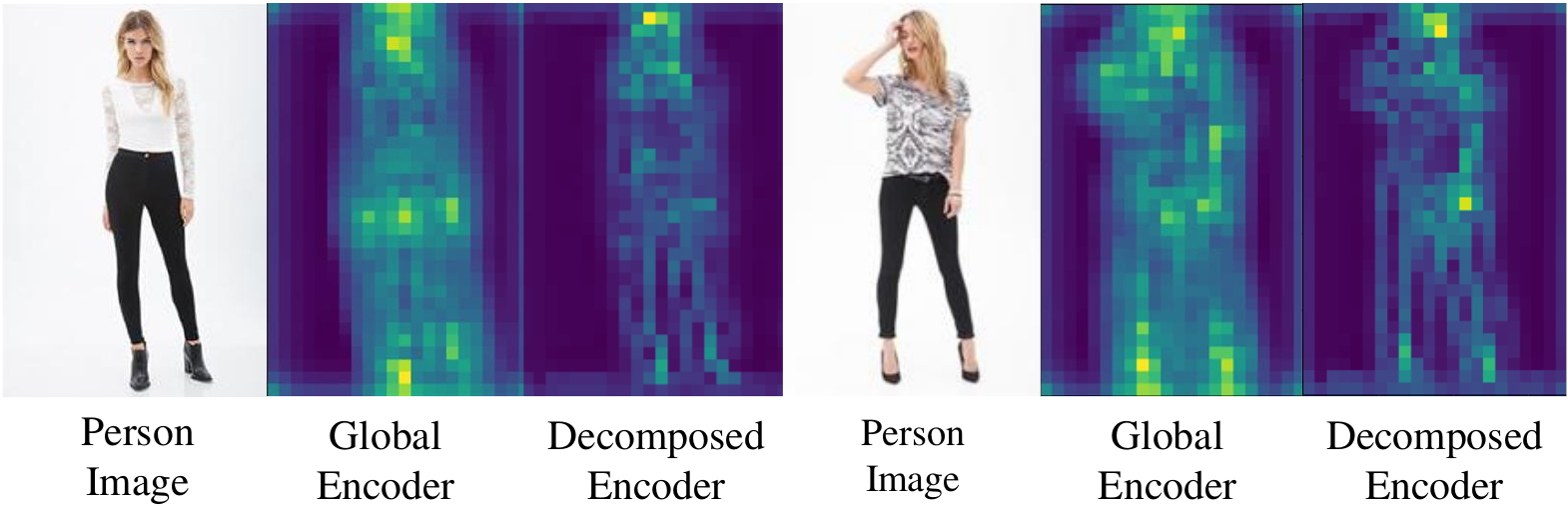}
    \caption{Feature map visualizations of global encoder and the DSE module. The structural information represented by the DSE module is significantly reduced compared with the global encoder.}
    \label{figure3}
\end{figure}

\subsection{Overall Workflow}

Without requiring unaligned data pairs, our method receives a single source image as input.
As shown in Figure \ref{figure2}, given a source person image $I$, we leverage the pre-trained human pose estimation model\cite{cao2017realtime} and human parser\cite{gong2017look} to obtain its pose skeleton $P$ and semantic mask $S$. 
For feature disentangling, we employ the pose encoder and the DSE module to extract the pose features $F_p\in\mathbb{R}^{C{\times}H{\times}W}$ and the semantic-aware decoupled style features $F_s\in\mathbb{R}^{C{\times}H{\times}W}$, respectively. 
For feature merging, the CMM module
is proposed to establish the dense spatial correlation field $\mathcal{C}$. 
Based on this correlation field, the $F_s$ could perform spatial rearrangement to obtain the merged features $F^*_s$. 
Eventually, the translation generator $G$ transforms the $F^*_s$ from the feature domain to realistic images. 

\subsection{Disentangled Feature Encoding}
For feature disentangling, there are two branches (\eg, pose branch, style branch) in our framework to encode pose features and style features, respectively.

\paragraph{Pose Encoding.} 
In the pose encoding branch, we employ the down-sampling convolutional neural networks (CNNs) to extract pose feature maps $F_p$ from the pose skeleton $P$.
Since the $F_p$ is globally encoded, its structure is aligned with source image $I$ inherently.

% The DSE module is designed to disentangle the semantic-aware appearances, which facilitates the person attribute editing. 
% Meanwhile, DSE can be viewed as an approach to create ``unaligned pairs " by generating style features that discard their structural information. 
% Different from \cite{zhang2020cross} which employed random geometric distortion to obtain unaligned data pairs, the DSE module could embed the person image from a complex manifold to the feature space according to different regions, which is conductive for the subsequent spatial correlation learning. 
% In this paper, inspired by the decomposed strategy in \cite{men2020controllable},
% %the decomposed component encoding strategy in
% we design the DSE module to obtain the semantic-aware disentangled style features, which can form `` unaligned data pairs " with their counterpart pose features. 
% Therefore, the source image itself could provide supervision information for the subsequent spatial correlation learning. 
% The feature map comparisons are shown in Fig.3.  
% We can observe that the signal strength distribution of the global encoder represents the structural information of the human body clearly, while the distribution of the DSE module is relatively flat, which indicates the structural information degradation. 

\begin{figure}[t]
\centering
\includegraphics[scale=0.3]{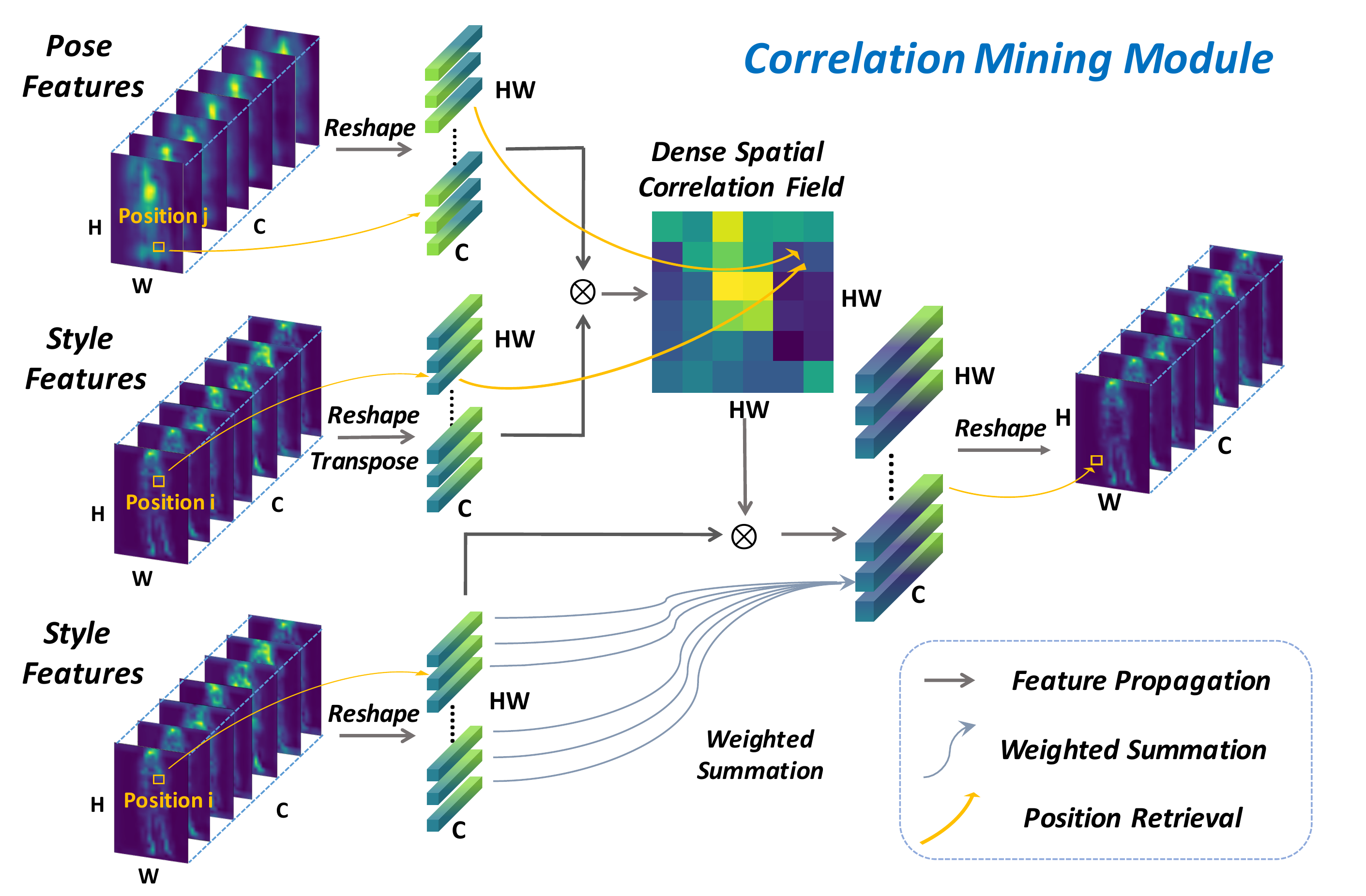}
\caption{Details of the Correlation Mining Module in our model. Each position of outputs is the weighted average summation of the input. The weights are stored in the correlation field.}
\label{figure4}
\end{figure}

% Specifically, 
% For self-supervised person image generation task, the key problem is how to extract the style features that are not aligned with its corresponding pose. Recently, \cite{men2020controllable} proposed decomposed component encoding (DCE) strategy to embed each semantic region of source image to style feature space separately, which disentangled the style feature from its pose and performed well in person attribute editing task. However, \cite{men2020controllable} transferred the statistics of style features to pose features and needed paired data to train. In this paper, we employ the DCE strategy to build a Decomposed style encoder for self-supervised training. 
\paragraph{Decomposed Style Encoding.}
For style encoding, we design the DSE module to obtain the semantic-aware decoupled style features $F_s$, which could form ``unaligned data pairs'' with their counterpart pose features.
Compared with the global encoder which encodes the person image entirely, the DSE module could embed the person image $I$ from a complex manifold to the feature space according to different regions.

As illustrated in Figure \ref{figure2}, we separate the segmentation map $S$ into 8-channel binary masks. 
Each channel indicates a specific body region (\eg, pants, hair). 
Then we employ element-wise multiplication between each binary mask and the source person image $I$ to obtain body parts.
In addition, we feed each body part into an encoder whose parameters are shared for all regions to extract the partial style features $F^i_s,i\in[1,8]$. 
Finally, we concatenate all $F^i_s$ along the channel dimension to construct the semantic-aware decoupled style features $F_s$.
Each position in style feature maps contains specific semantic information. 
Furthermore, for eliminating the limitation caused by the fixed concatenation order, we propose a Cross Channel Fusion (CCF) module to endow plentiful information into each position by selecting desired semantic features from different semantic regions. 
In structure, the CCF module has a concise design which consists of two 1$\times$1 convolutional blocks.

To verify the effect of DSE, we visualize the feature maps extracted by the global encoder and the DSE, respectively.
As shown in Figure \ref{figure3}, we can observe that the signal strength distribution of the global encoder represents the structural information clearly, while the distribution of the DSE is relatively flat, which indicates the structural information degradation.

% to obtain each semantic region $I^i_s$ of source images. Then each semantic region is feed into style encoder to extract per-region style feature map. Finally, each feature map is concatenated to construct a semantic-decoupled style feature map.

\subsection{Correlation based Feature Merging}%名词
In the merging phase, we propose the CMM module, which aims to establish the dense spatial correlation field $\mathcal{C}$ for feature rearrangement. 
To begin with, we reshape the feature $F_i$ into $[F_i(1),F_i(2),\cdots,F_i(hw)]\in\mathbb{R}^{C{\times}HW},i\in\left\{p,s\right\}$. 
Each vector $F_i(j)\in\mathbb{R}^C$ in $F_i$ represents the semantic information of the $j^{th}$ location in the feature map, $j\in[1,{hw}]$.

As illustrated in Figure \ref{figure4}, given $F_s$ and $F_p$, each vector $F_s(i)$ serves as the query to retrieve the relevant key $F_p(j)$ from the $F_p$. 
Therefore, the correlation field $\mathcal{C}\in\mathbb{R}^{HW{\times}HW}$ is established, whose element $C_{ij}$ is the correlation of key-value pairs, followed by a softmax activation. 
\begin{equation}
  C_{ij} = \frac{exp(s_{ij})}{\sum_{i=1}^{hw}exp(s_{ij})} 
  \label{eq:also-important}
\end{equation}

\begin{equation}
  s_{ij} = \frac{\bar{F}_s(i)\bar{F}_p(j)}{||\bar{F}_s(i)|| \ ||\bar{F}_p(j)||}
  \label{eq:also-important}
\end{equation}
where $\bar{F}_s(i)$ and $\bar{F}_p(j)$ represent the centralized feature, \ie $\bar{F}_s(i)$ = $F_s(i)$ - mean($F_s(i)$). 
The correlation field $\mathcal{C}$ contains the weights which could be assigned to value vectors for feature rearrangement.
Specifically, the rearranged feature $F^*_s = [F^*_s(1), F^*_s(2),\cdots, F^*_s(hw)]\in\mathbb{R}^{C{\times}HW}$ is obtained by calculating the weighted average summation of all positions in
feature $F_s$. 

\begin{equation}
  F^*_s(i) = \sum_{j=1}^{hw}c_{ij}F_s(j), \\\\\\i\in[1,hw]
  \label{eq:also-important}
\end{equation}

Based on the above operations, the $F^*_s$ is structurally aligned with the input pose which could be fed into the translation generator to synthesize a realistic person image.

% Then we warp the style features, in other words, the value vectors $F_s$ into the pose features according to the established correlation field $\mathcal{C}$ by calculating the weighted average of all positions of $F_s$. 
% The warped style feature is $F^*_s = [F^*_s(1), F^*_s(2),\cdots, F^*_s(hw)]\in\mathbb{R}^{C{\times}HW}$, in which
% \begin{equation}
%   F^*_s(i) = \sum_{j=1}^{hw}c_{ij}F_s(j), \\\\\\i\in[1,hw]
%   \label{eq:also-important}
% \end{equation}

% Based on the above operations, the warped style feature $F^*_s$ serves as a coarse result whose structure is aligned with the input pose. 
% Finally, the $F^*_s$ will be fed into the translation generator to synthesize a realistic image.

%-------------------------------------------------------------------------
\subsection{Aligned Feature Translation}
With the rearranged features $F^*_s$ as input, the translation generator could synthesize the target image $\hat{I}$ for self-reconstruction. 
To better preserve the structural information, we employ the U-Net architecture\cite{ronneberger2015u} as our translation generator, as its skip connection propagates the information directly from encoder to decoder.

\subsection{Objective Functions}
\paragraph{Adversarial Learning.} Following the configuration of \cite{zhu2019progressive}, we employ two discriminators, one is a pose discriminator $D_p$ to maintain the pose consistency, and the other is a style discriminator $D_s$ to maintain the style consistency. Both of them promote the generator $G$ to generate realistic images.
The adversarial loss $L_{adv}$ is listed as follows:
% \begin{equation}
\begin{align}
    \mathcal{L}_{adv}& = \mathbb{E}_{I,P}[\log(D_s(I)\cdot D_p(I,P))]
    \notag
    \\&+\mathbb{E}_{I,P}[\log((1-D_s(G(I,P))) 
    \notag
    \\&\cdot(1-D_p(G(I,P))))] 
  \label{eq:also-important}
\end{align}
% \end{equation}

% \begin{align}
% \mathcal{L}_{total} = \mathcal{L} _{GAN}&+\alpha \mathcal{L}_{content}+\lambda \mathcal{L} _{AR}
% \notag
% \\&+\delta \mathcal{L} _{perceptual}+\eta \mathcal{L} _{BCE}.
% \end{align}

% \begin{align}
% \mathcal{L}_{total} = \mathcal{L} _{GAN}&+\alpha \mathcal{L}_{content}+\lambda \mathcal{L} _{perceptual}
% \notag
% \\&+\delta \mathcal{L} _{BCE}+\eta \mathcal{L} _{IAG}
% \notag
% \\&+\tau \mathcal{L} _{ITG}+\xi \mathcal{L} _{ICT}.
% \end{align}

\paragraph{Self-supervised Reconstruction.} The reconstruction loss $L_{rec}$ can be formulated as the L1 distance between the source image $I$ and generated image $\hat{I}$, which encourages the $\hat{I}$ to be similar with the $I$ at the pixel level. 
\begin{equation}
\begin{split}
    \mathcal{L}_{rec} = {||  \hat{I}-I  ||}_1
  \label{eq:also-important}
\end{split}
\end{equation}

\begin{figure}[t]
    \centering
    \includegraphics[scale=0.5]{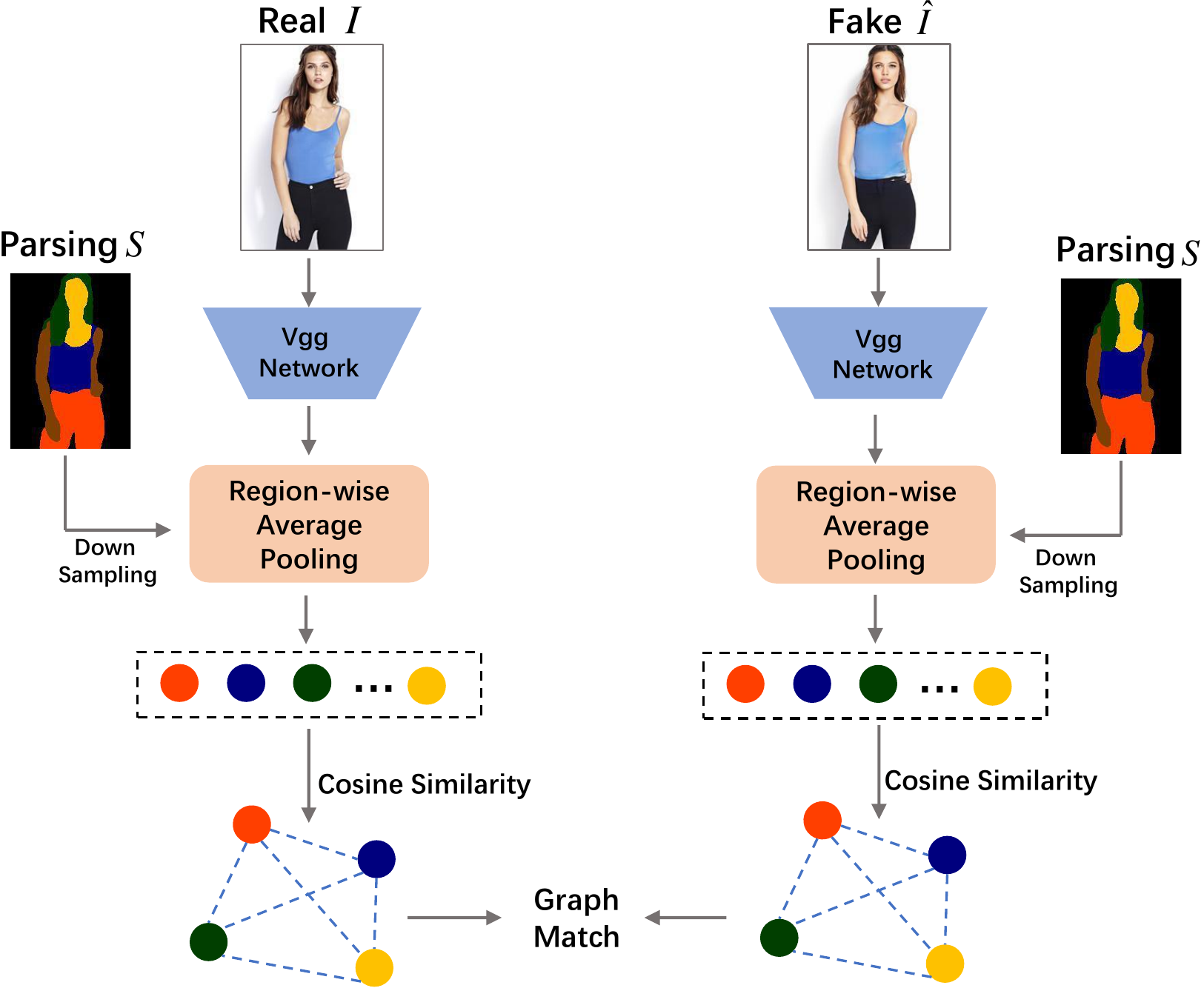}
    \caption{Details of the graph generator in our model. Nodes represent per-region styles and the edges measure the similarities between nodes.}
    \label{figure5}
\end{figure}

\paragraph{Perceptual Consistency.} The perceptual loss $L_{perc}$ calculates the $L_1$ distance between the pre-trained VGG features of $I$ and $\hat{I}$, which measures the high-level semantic differences between images \cite{johnson2016perceptual}.
\begin{equation}
\begin{split}
    \mathcal{L}_{perc} = {||\phi^l(\hat{I})-  \phi^l(I)  ||}_1
  \label{eq:also-important}
\end{split}
\end{equation}

\paragraph{Style Consistency.} The style loss $L_{style}$ calculates the statistical errors between the pre-trained VGG features of $I$ and $\hat{I}$, which penalizes the difference in colors and textures \cite{johnson2016perceptual}. As shown in Formula (7), $\phi^l$ is the activation at the $j$th layer of the pre-trained VGG network, and $\mathbb{G} $ is the Gram matrix.
\begin{equation}
\begin{split}
    \mathcal{L}_{style} = \sum_{l}{||\mathbb{G}(\phi^l(\hat{I})) - \mathbb{G}(\phi^l(I))||}_1
  \label{eq:also-important}
\end{split}
\end{equation}

% \textbf{Correlation loss.}  Previous spatial correlation learning methods usually lack direct supervision for establishing the correlation field \cite{he2018deep,zhang2019deep}, which makes it difficult to ensure that the model learns meaningful correlation. To this end, the correlation loss is designed to guarantee the warped style features $F_s^*$ and the perceptual feature of target images are aligned in the same domain. The specific formula is as follows,
% \begin{equation}
% \begin{split}
%     \mathcal{L}_{cor} = {||F_s^* -\phi^l(I)||}_1
%   \label{eq:also-important}
% \end{split}
% \end{equation}

\begin{figure*}[ht]
    \centering
    \includegraphics[scale=0.50]{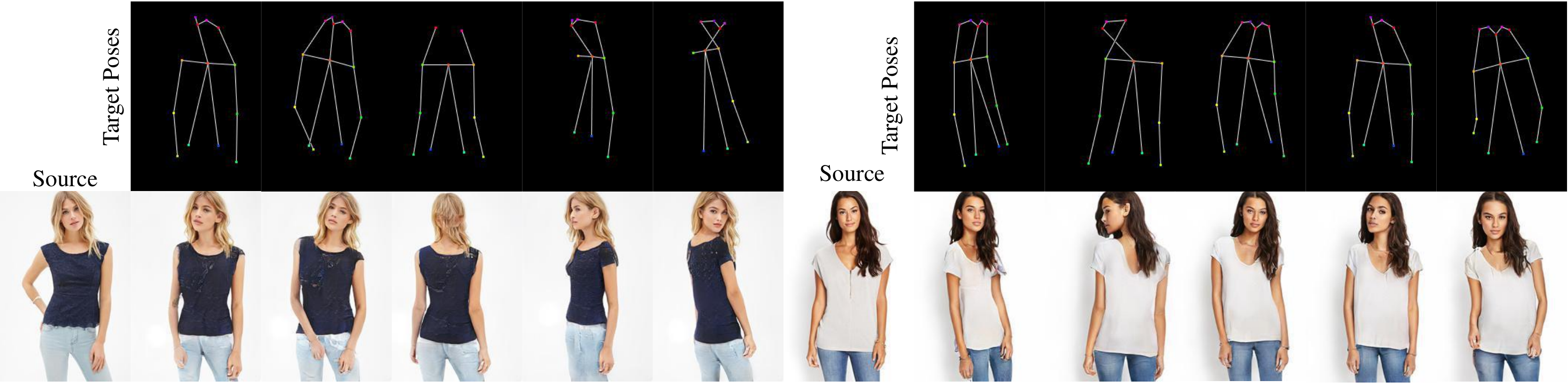}
    \caption{The results of our method in the pose guided person image generation.}
    \label{figure6posetransfer}
\end{figure*}

\paragraph{Body Structure Retaining.} 
% Due to the training process is self-supervised with a single pose in each iteration,
% %by the reconstruction of source images, 
% the model cannot make a reasonable completion for the unknown regions when performing cross-scale pose transformation, especially half body to full body. 
% To this end, we propose the Body Structure Retaining Loss to help the network capture more body structure information. 
The BSR Loss is proposed to endow prior knowledge of invisible regions through constraining semantic relationships among body parts. 
We design a graph generator to model this relationship. 
As illustrated in Figure \ref{figure5}, we employ a pre-trained VGG network and the region-wise average pooling layer\cite{zhu2020sean} to obtain the body graph $\mathbb{M}$, in which the nodes represent per-region styles and the edges measure the similarities between nodes. 

Due to the training process is self-supervised with the single pose in each iteration, the model cannot make a reasonable completion for the unknown regions when performing cross-scale pose transformation.
Applying BSR Loss for training encourages the output person image to retain a reasonable structure, which is conductive for half body to full body transformation. 
% Specifically, we design a graph generator to extract the graph representation $\mathbb{M}$ of person images, as illustrated in Fig.5. 
% Each node of the graph corresponds to the perceptual features of a semantic region and each edge measures the similarity between each node pair. 
% Therefore, the network learns the region features and their correlations simultaneously.
% When the model completes the lower part of the body, the contextual correlation will be taken into consideration to generate a reasonable full body image. 
We calculate the BSR loss between $I$ and $\hat{I}$ as the $L_{graph}$. 
\begin{equation}
\begin{split}
    \mathcal{L}_{graph} = {||\mathbb{M}(I,S)-\mathbb{M}(\hat{I},S)||}_1
  \label{eq:also-important}
\end{split}
\end{equation}

The overall objective function is shown in Formula (9), where $\alpha_{adv}$, $\alpha_{rec}$, $\alpha_{perc}$, $\alpha_{style}$, $\alpha_{garph}$ are the weights of the corresponding loss functions.
%%%%这些参数记得Expriments提一下，别忘了。。。
% \begin{align}
%     \mathcal{L}_{total}& = \alpha_{adv}\mathcal{L}_{adv}+\alpha_{rec}\mathcal{L}_{rec}\\&
%     \notag
%     +\alpha_{perc}\mathcal{L} _{perc}+\alpha_{style}\mathcal{L} _{style}\\&
%     \notag
%     +\alpha_{cor}\mathcal{L} _{cor}+\alpha_{garph}\mathcal{L} _{garph}
%     \notag
%   \label{eq:also-important}
% \end{align}

\begin{align}
    \mathcal{L}_{total}& = \alpha_{adv}\mathcal{L}_{adv}+\alpha_{rec}\mathcal{L}_{rec}+\alpha_{perc}\mathcal{L} _{perc}\\&
    \notag
    +\alpha_{style}\mathcal{L} _{style}+\alpha_{graph}\mathcal{L}_{graph}\\&
    \notag
  \label{eq:also-important}
\end{align}

\section{Experiments}
\subsection{Implementation Details}
\paragraph{Dataset.} We carry out our experiments on DeepFashion In-shop Clothes Retrieval Benchmark \cite{liu2016deepfashion}, which contains 52,712 high quality person images.
%with clean backgrounds. 
We split the dataset following the same configurations of \cite{ma2021must}, 

\paragraph{Metrics.} 
We use the common metrics such as Structural Similarity (SSIM)\cite{wang2004image}, Inception Score (IS)\cite{salimans2016improved}, Learned Perceptual Image Patch Similarity (LPIPS)\cite{zhang2018unreasonable}, and Fréchet Inception Distance (FID)\cite{heusel2017gans} to assess the quality of generated images quantitatively. 
SSIM indicate the similarity between paired images in raw pixel space. 
% Meanwhile, LPIPS, IS and FID calculate the perceptual distance between the generated images and ground truth images in the feature space.
Meanwhile, LPIPS, IS and FID measure the realism of the generated images at the feature level.
\paragraph{Network Architecture and Training Details.} 
Both the pose encoder and style encoder employ several downsampling convolutional layers to extract features. 
The feature maps with resolutions $32\times32$ are applied for establishing the correlation field. 
Our method is implemented on PyTorch framework using 4 Nvidia TitanX GPUs. 
The weights for loss functions are set to $\alpha_{adv}=5$, $\alpha_{rec}=1$, $\alpha_{perc}=1$, $\alpha_{style}=150$, $\alpha_{garph}=1$, respectively. 
%%%记得写上超参数等于什么。。。

%We adopt the Adam optimizer \cite{kingma2014adam} to train our model. %sun

\begin{figure}[t]
    \centering
    \includegraphics[scale=0.33]{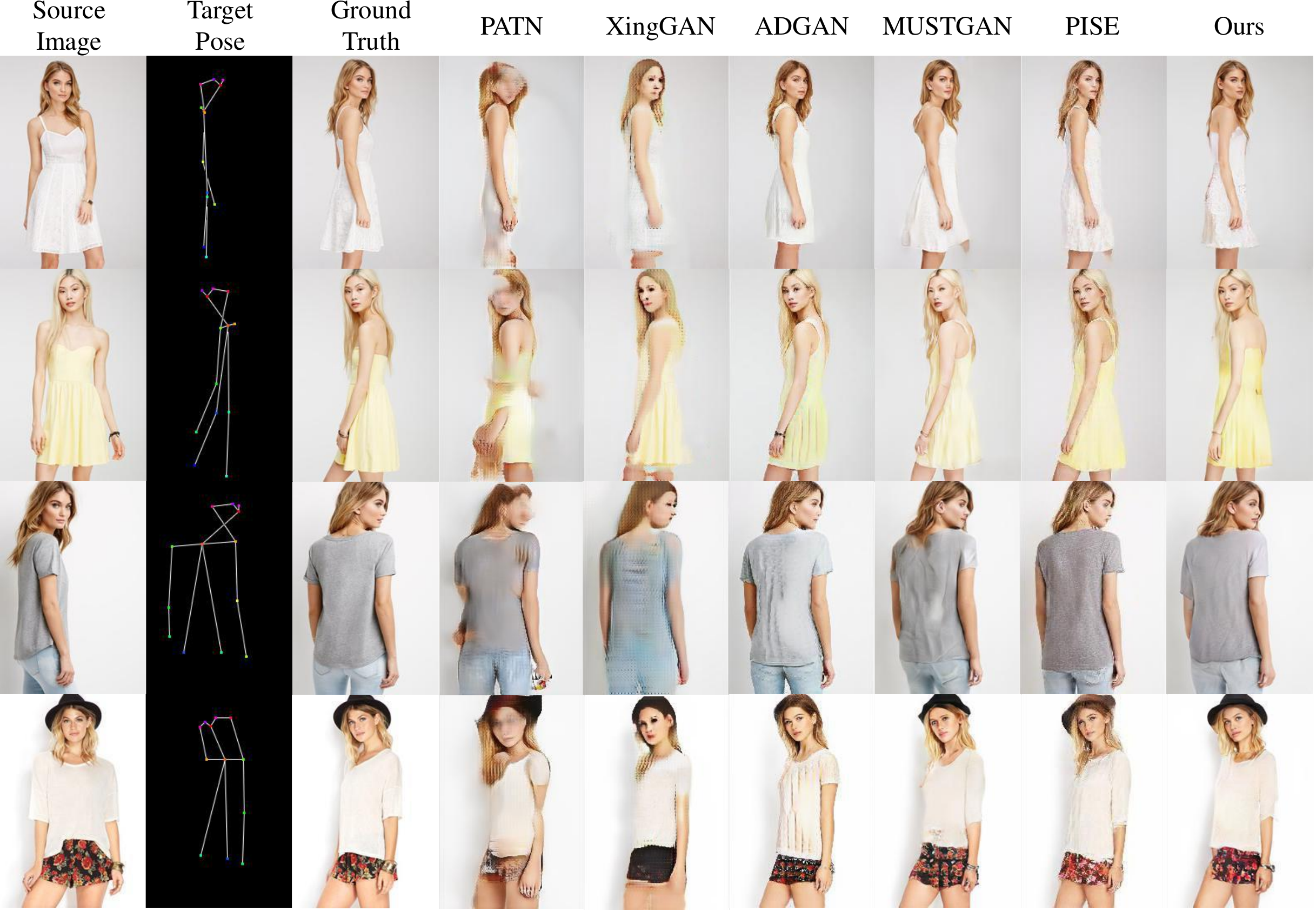}
    \caption{The comparisons with other state-of-the-art methods on pose guided person image generation. Zoom in for a better view.}
    \label{figure7compare}
\end{figure}

\subsection{Pose Guided Person Image Generation}
Pose guided person image generation, or pose transfer, aims to change the posture of a person image given target poses as condition.
Pose transfer is an important application of person image generation. 
As shown in Figure \ref{figure1} (left) and Figure \ref{figure6posetransfer} (all), given a source person image, our model could transform it to any target pose and keep the appearance details unchanged.
% \subsubsection{}

\paragraph{Qualitative Comparison.} 
We compare the generated images of our method with several state-of-the-art approaches, including PATN\cite{zhu2019progressive}, XingGAN\cite{tang2020xinggan}, ADGAN\cite{men2020controllable}, MUSTGAN\cite{ma2021must} and PISE\cite{zhang2021pise}. 
All the results are obtained using the source code or the pre-trained model released by the authors. 
The results of the qualitative comparisons are shown in Figure \ref{figure7compare}. PATN and XingGAN generate blurry results since these models can not disentangle different features. The results of ADGAN and MUSTGAN have correct postures, but they fail to maintain detailed textures. This is because these models can not capture the spatial correlation well. 
PISE could generate desirable results. 
However, its results still have some unsatisfactory artifacts due to the lack of semantic relationships. Meanwhile, this model requires unaligned image pairs for training. In contrast, our model obtain competitive results only requires source images.
%Our method could... 
%More results can be found in Supp. %sun

\paragraph{Quantitative Comparison.} 
As shown in Table \ref{table1}, we compare our method with several state-of-the-art supervised and unsupervised methods on the DeepFashion.
As we can see, our method outperforms these methods in most metrics on both supervised and unsupervised setting, which demonstrates the superiority of our method in generating high-quality person images. 
%``-" denotes that the performance cannot be acquired according to the original paper.

\begin{figure*}[ht]
    \centering
    \includegraphics[scale=0.50]{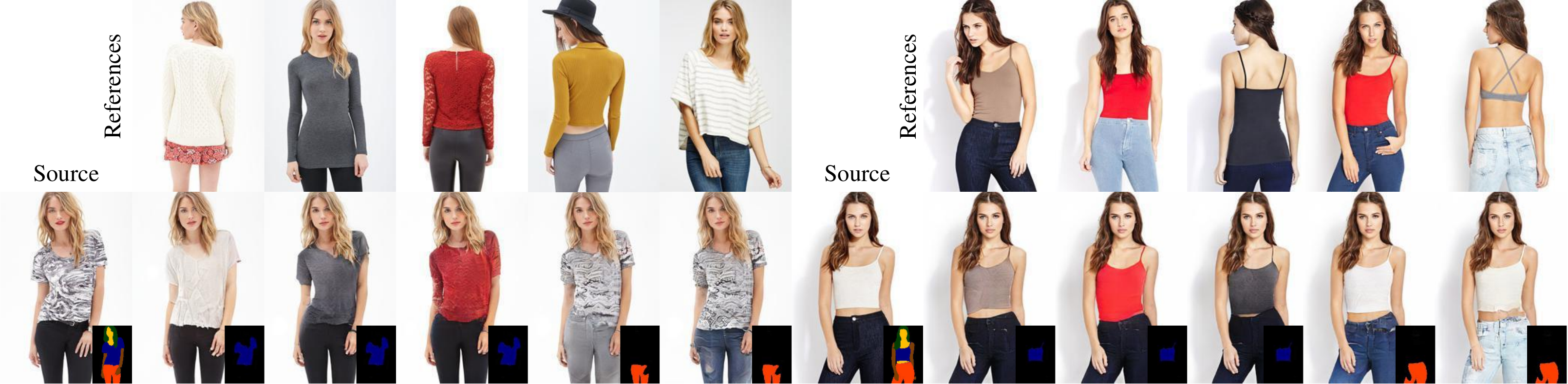}
    \caption{The results of our method in the person attribute editing.}
    \label{figure8personedit}
\end{figure*}

\begin{table}[ht]
\centering
\caption{Quantitative comparisons with other supervised and unsupervised methods on DeepFashion.}
\begin{tabular}{lllll}
\hline
Method   & FID$\downarrow$      & SSIM$\uparrow$   & LPIPS$\downarrow$      & IS$\uparrow$     \\\hline
\textbf{\emph{Unsupervised}} &     &      &      &     \\
% UPIS\cite{pumarola2018unsupervised}         & -        & 0.747  & -          & 2.97   \\
VU-Net\cite{esser2018variational}           & 23.583   & 0.786  & 0.3211     & 3.087  \\
E2E\cite{song2019unsupervised}              & 29.9     & 0.736  & 0.238      & 3.441  \\
DPIG\cite{ma2018disentangled}               & 48.2     & 0.614  & 0.284      & 3.228  \\
MUST\cite{ma2021must}                       & 15.902   & 0.742  & -          & 3.692  \\\hline
\textbf{\emph{Supervised}}   &     &      &      &     \\
Intr-Flow\cite{li2019dense}                 &  16.134  & 0.798  & 0.2131     & 3.251  \\
Def-GAN\cite{siarohin2018deformable}        &  18.547  & 0.770  & 0.2994     & 3.141  \\
PATN\cite{zhu2019progressive}               &  24.071  & 0.770  & 0.2520     & 3.213  \\
ADGAN\cite{men2020controllable}             &  18.395  & 0.771  & 0.2242     & 3.329  \\
GFLA\cite{ren2020deep}                      &  14.061  & 0.701  & 0.2219     & 3.635  \\
PISE\cite{zhang2021pise}                    &  13.61   &   -    & 0.2059     &  -     \\\hline
SCM-Net                                  &  \textbf{12.18}   & 0.751  & 0.2142    & 3.632 \\\hline
\end{tabular}
\label{table1}
\end{table}

\begin{table}[ht]
\centering
\caption{The evaluation results of ablation study.}
\begin{tabular}{lllll}
\hline
Method       & FID$\downarrow$      & SSIM$\uparrow$   & LPIPS$\downarrow$      & IS$\uparrow$         \\ \hline
w/o DSE      & 12.86    & 0.750  & 0.187  & 3.2456  \\
w/o CCF      & 17.08    & 0.751  & 0.175  & 3.605   \\
w/o BSR      & 12.61  & 0.755  & 0.178  & 3.441    \\
Full         &\textbf{12.18}    & 0.751  & 0.182  &\textbf{3.632}   \\\hline
\end{tabular}
\label{table2}
\end{table}

\subsection{Ablation Study}
We further perform the ablation study to analyze the contribution of each module and the proposed BSR Loss in our method. 
Firstly, we introduce the variants implemented by alternatively removing a specific component from our full model. There are four settings in this module ablation.
1). \textbf{W/o DSE.} This model removes the DSE module and directly uses a global encoder to extract the style features.
2). \textbf{W/o CCF.} This model removes the Cross Channel Fusion module from the DSE.
3). \textbf{W/o BSR.} This model removes the BSR Loss during the training procedure.
4). \textbf{Full.} This model represents our full model.

\begin{figure}[t]
    \centering
    \includegraphics[scale=0.40]{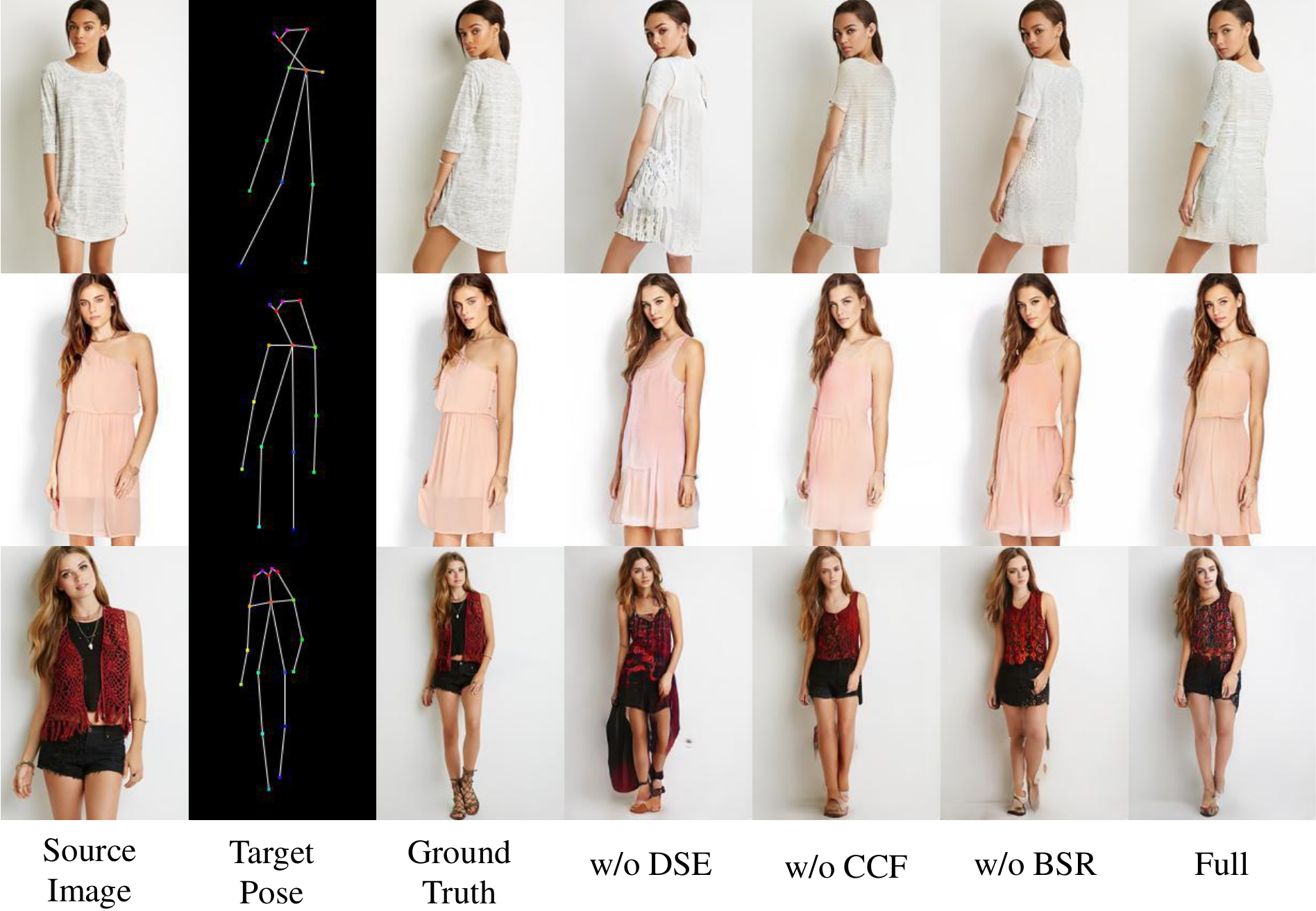}
    \caption{The qualitative comparison of ablation study.}
    \label{figure9ablationstudy}
\end{figure}

\begin{figure}[t]
    \centering
    \includegraphics[scale=0.45]{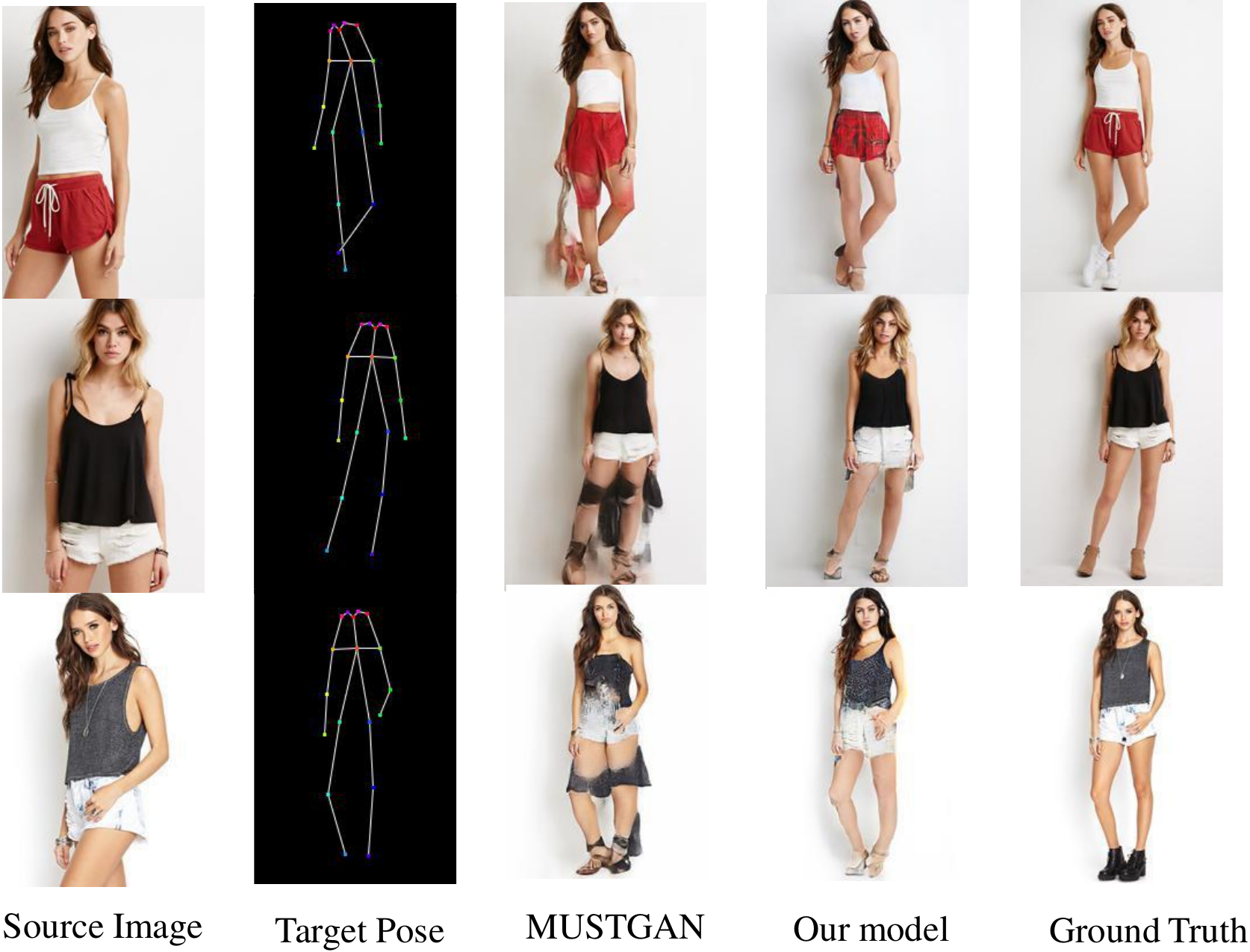}
    \caption{Results of half body to full body transformation}
    \label{figure10halftofull}
\end{figure}

Table \ref{table2} shows the quantitative results of the ablation study. 
We can observe that our full model achieves the best performance on FID and IS metrics. 
Meanwhile, the removal of any components will degrade the performance of the model integrally.
Figure \ref{figure9ablationstudy} shows the qualitative comparisons of different ablation models. 
We can observe that the w/o DSE model fails to preserve the styles of source images and the w/o CCF model has limitation in preserving the detailed texture. 
Meanwhile, w/o BSR model can not complete the lower body well while the full model could generate reasonable results. 
It demonstrates that BSR Loss enhance the model's ability of capturing body structural information.
Furthermore, we illustrate the comparisons on half body to full body transformation with previous self-supervised method MUST-GAN\cite{ma2021must}. 
Figure \ref{figure10halftofull} shows the advantages of our method when performing half body to full body transformation. 
We can observe that MUST-GAN\cite{ma2021must} would generate more artifacts, while our method could complete the lower part of the body reasonably with correlation learning. 

\subsection{Person Attribute Editing}
Our model can also achieve person attribute editing based on reference images by exchanging channel features of specific semantic areas in semantic-aware decoupled style features. 
As shown in Figure \ref{figure1} (right) and Figure \ref{figure8personedit} (all), our method could edit the style of the upper clothes, pants and hair style respectively.

\subsection{Applications on Face Generation tasks}
In this section, we demonstrate the versatility of our method. 
Since our method could disentangle the shape and style features, it could also be applied to other image generation tasks under this self-supervised framework. 
Two face generation tasks are shown as follows.

\begin{figure}[t]
    \centering
    \includegraphics[scale=0.48]{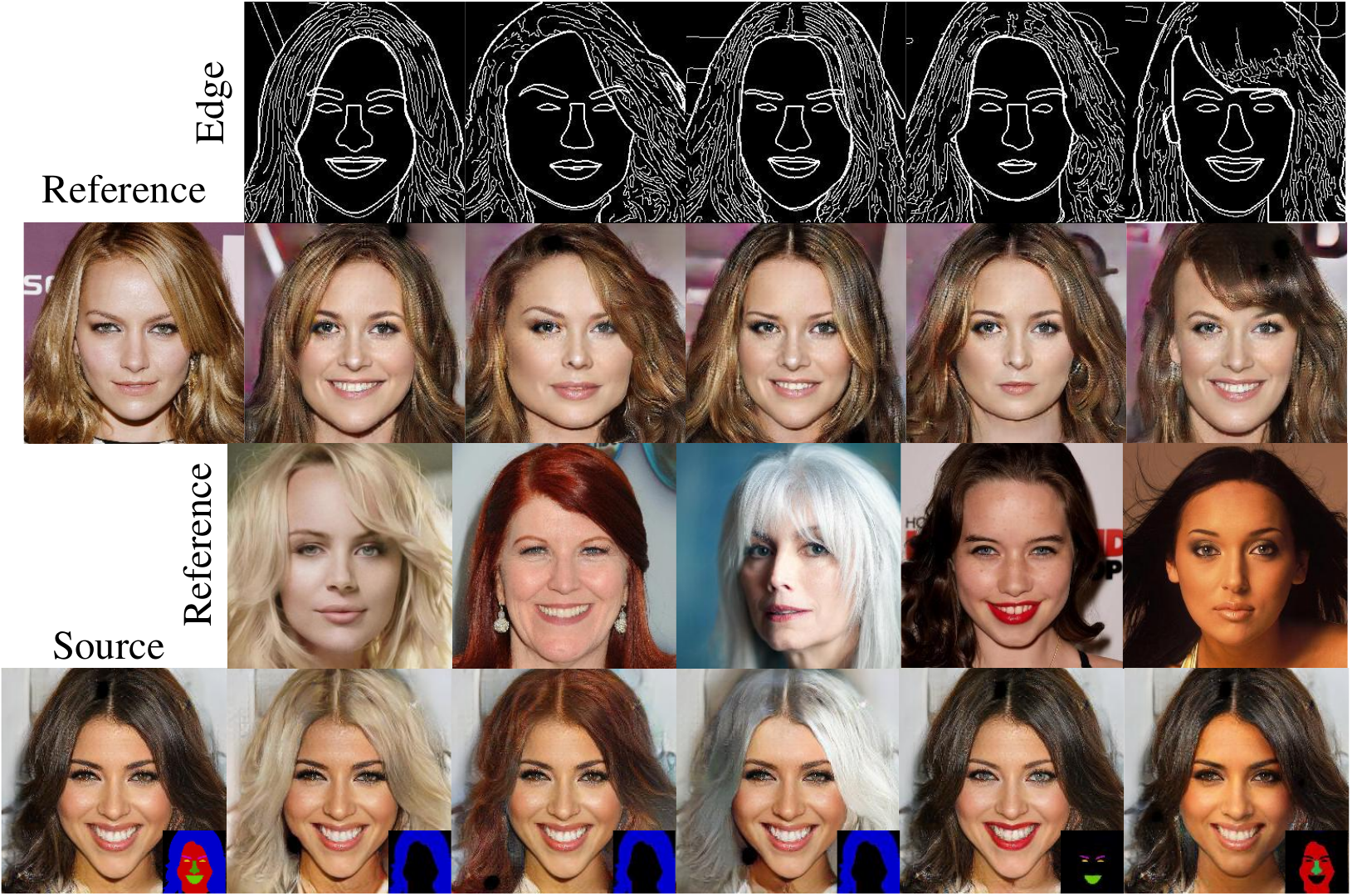}
    \caption{The results of our method in reference-based face edge colorization (top) and face attribute editing (bottom).}
    \label{figure11edge2face}
\end{figure}

\paragraph{Reference-Based Edge Colorization.} 
Reference-based edge colorization aims to translate an edge map to a realistic image based on a reference image. 
Regarding the edge map as a pose skeleton and the reference image as a person image, our self-supervised model could achieve edge colorization. 
We obtain the edge map following \cite{zhang2020cross} and use CelebA-HQ\cite{karras2017progressive} dataset for training. 
The results are shown in Figure \ref{figure11edge2face} (top). We can observe that the results maintain a good style consistency with the reference image, and preserve a good shape consistency with input edge maps.

\paragraph{Face Attribute Editing.} Similar to person attribute editing, our method could also achieve face attribute editing. The results can be found in Figure \ref{figure11edge2face} (bottom). We can edit specific attributes while keeping other attributes unchanged.

% \begin{figure}[t]
%     \centering
%     \includegraphics[scale=0.48]{latex/figure/figure12-cropped.pdf}
%     \caption{The results of our method in face attribute editing.}
%     \label{figure12faceediting}
% \end{figure}

\section{Limitation}
\begin{figure}[h]
    \centering
    \includegraphics[scale=0.5]{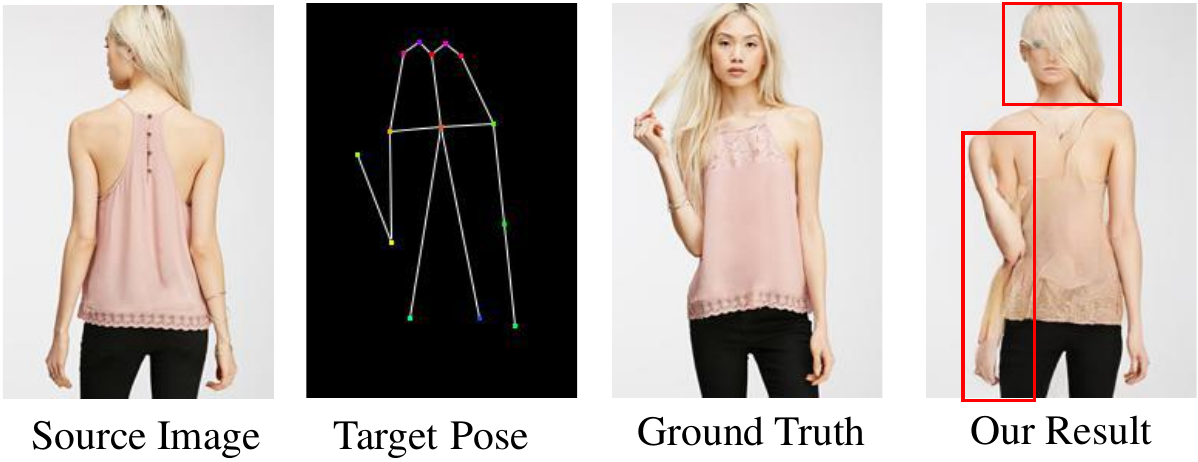}
    \caption{The illustrations of the model limitation. The hair and left arm are transferred directly from source image.}
    \label{limitation}
\end{figure}

% Due to the training process is promoted by the reconstruction of source images, the model will remember some source patterns when performing pose transfer. 
% As shown in Figure \ref{limitation}, the hair and left arm are directly copied from source image. This, we hypothesis, is caused by insufficient disentanglement of style and pose features. 
% 
As shown in figure \ref{limitation}, our self-supervised model sometimes directly transfers certain source patterns into the final results when performing pose transfer, which is a rare situation in the supervised model. 
This, we hypothesis, is caused by the inherent defects of self-supervised strategy that the self-reconstruction process makes the model easy to overfit.
This phenomenon might be avoided by employing spatial transformation to perform data augmentation during training in the future.

% 孙注释掉了这一部分

\section{Conclusion}
In this paper, we propose a Self-supervised Correlation Mining Network (SCM-Net) for person image generation. 
We propose two specially designed modules, the DSE module for feature disentanglement, and the CMM module for feature merging based on the spatial correlation. 
Meanwhile, the BSR Loss is proposed to promote our network to better capture the structural information, especially for half body to full body transformation. 
Extensive experiment results conducted on person and face datasets demonstrate the superiority of our method.

%%%%%%%%% REFERENCES
{\small
\bibliographystyle{ieee_fullname}
\bibliography{main}

\begin{thebibliography}{10}\itemsep=-1pt

\bibitem{cao2016deep}
Shaosheng Cao, Wei Lu, and Qiongkai Xu.
\newblock Deep neural networks for learning graph representations.
\newblock In {\em Proceedings of the AAAI Conference on Artificial
  Intelligence}, volume~30, 2016.

\bibitem{cao2017realtime}
Zhe Cao, Tomas Simon, Shih-En Wei, and Yaser Sheikh.
\newblock Realtime multi-person 2d pose estimation using part affinity fields.
\newblock In {\em Proceedings of the IEEE conference on computer vision and
  pattern recognition}, pages 7291--7299, 2017.

\bibitem{dong2018soft}
Haoye Dong, Xiaodan Liang, Ke Gong, Hanjiang Lai, Jia Zhu, and Jian Yin.
\newblock Soft-gated warping-gan for pose-guided person image synthesis.
\newblock {\em arXiv preprint arXiv:1810.11610}, 2018.

\bibitem{esser2018variational}
Patrick Esser, Ekaterina Sutter, and Bj{\"o}rn Ommer.
\newblock A variational u-net for conditional appearance and shape generation.
\newblock In {\em Proceedings of the IEEE Conference on Computer Vision and
  Pattern Recognition}, pages 8857--8866, 2018.

\bibitem{gong2017look}
Ke Gong, Xiaodan Liang, Dongyu Zhang, Xiaohui Shen, and Liang Lin.
\newblock Look into person: Self-supervised structure-sensitive learning and a
  new benchmark for human parsing.
\newblock In {\em Proceedings of the IEEE Conference on Computer Vision and
  Pattern Recognition}, pages 932--940, 2017.

\bibitem{goodfellow2014generative}
Ian Goodfellow, Jean Pouget-Abadie, Mehdi Mirza, Bing Xu, David Warde-Farley,
  Sherjil Ozair, Aaron Courville, and Yoshua Bengio.
\newblock Generative adversarial nets.
\newblock {\em Advances in neural information processing systems}, 27, 2014.

\bibitem{han2019clothflow}
Xintong Han, Xiaojun Hu, Weilin Huang, and Matthew~R Scott.
\newblock Clothflow: A flow-based model for clothed person generation.
\newblock In {\em Proceedings of the IEEE/CVF International Conference on
  Computer Vision}, pages 10471--10480, 2019.

\bibitem{he2018deep}
Mingming He, Dongdong Chen, Jing Liao, Pedro~V Sander, and Lu Yuan.
\newblock Deep exemplar-based colorization.
\newblock {\em ACM Transactions on Graphics (TOG)}, 37(4):1--16, 2018.

\bibitem{heusel2017gans}
Martin Heusel, Hubert Ramsauer, Thomas Unterthiner, Bernhard Nessler, and Sepp
  Hochreiter.
\newblock Gans trained by a two time-scale update rule converge to a local nash
  equilibrium.
\newblock {\em Advances in neural information processing systems}, 30, 2017.

\bibitem{hou2020inter}
Yuenan Hou, Zheng Ma, Chunxiao Liu, Tak-Wai Hui, and Chen~Change Loy.
\newblock Inter-region affinity distillation for road marking segmentation.
\newblock In {\em Proceedings of the IEEE/CVF Conference on Computer Vision and
  Pattern Recognition}, pages 12486--12495, 2020.

\bibitem{huang2017arbitrary}
Xun Huang and Serge Belongie.
\newblock Arbitrary style transfer in real-time with adaptive instance
  normalization.
\newblock In {\em Proceedings of the IEEE International Conference on Computer
  Vision}, pages 1501--1510, 2017.

\bibitem{johnson2016perceptual}
Justin Johnson, Alexandre Alahi, and Li Fei-Fei.
\newblock Perceptual losses for real-time style transfer and super-resolution.
\newblock In {\em European conference on computer vision}, pages 694--711.
  Springer, 2016.

\bibitem{karras2017progressive}
Tero Karras, Timo Aila, Samuli Laine, and Jaakko Lehtinen.
\newblock Progressive growing of gans for improved quality, stability, and
  variation.
\newblock {\em arXiv preprint arXiv:1710.10196}, 2017.

\bibitem{lee2020reference}
Junsoo Lee, Eungyeup Kim, Yunsung Lee, Dongjun Kim, Jaehyuk Chang, and Jaegul
  Choo.
\newblock Reference-based sketch image colorization using augmented-self
  reference and dense semantic correspondence.
\newblock In {\em Proceedings of the IEEE/CVF Conference on Computer Vision and
  Pattern Recognition}, pages 5801--5810, 2020.

\bibitem{li2019dense}
Yining Li, Chen Huang, and Chen~Change Loy.
\newblock Dense intrinsic appearance flow for human pose transfer.
\newblock In {\em Proceedings of the IEEE/CVF Conference on Computer Vision and
  Pattern Recognition}, pages 3693--3702, 2019.

\bibitem{liao2017visual}
Jing Liao, Yuan Yao, Lu Yuan, Gang Hua, and Sing~Bing Kang.
\newblock Visual attribute transfer through deep image analogy.
\newblock {\em arXiv preprint arXiv:1705.01088}, 2017.

\bibitem{liu2016deepfashion}
Ziwei Liu, Ping Luo, Shi Qiu, Xiaogang Wang, and Xiaoou Tang.
\newblock Deepfashion: Powering robust clothes recognition and retrieval with
  rich annotations.
\newblock In {\em Proceedings of the IEEE conference on computer vision and
  pattern recognition}, pages 1096--1104, 2016.

\bibitem{ma2017pose}
Liqian Ma, Xu Jia, Qianru Sun, Bernt Schiele, Tinne Tuytelaars, and Luc
  Van~Gool.
\newblock Pose guided person image generation.
\newblock {\em arXiv preprint arXiv:1705.09368}, 2017.

\bibitem{ma2018disentangled}
Liqian Ma, Qianru Sun, Stamatios Georgoulis, Luc Van~Gool, Bernt Schiele, and
  Mario Fritz.
\newblock Disentangled person image generation.
\newblock In {\em Proceedings of the IEEE Conference on Computer Vision and
  Pattern Recognition}, pages 99--108, 2018.

\bibitem{ma2021must}
Tianxiang Ma, Bo Peng, Wei Wang, and Jing Dong.
\newblock Must-gan: Multi-level statistics transfer for self-driven person
  image generation.
\newblock In {\em Proceedings of the IEEE/CVF Conference on Computer Vision and
  Pattern Recognition}, pages 13622--13631, 2021.

\bibitem{men2020controllable}
Yifang Men, Yiming Mao, Yuning Jiang, Wei-Ying Ma, and Zhouhui Lian.
\newblock Controllable person image synthesis with attribute-decomposed gan.
\newblock In {\em Proceedings of the IEEE/CVF Conference on Computer Vision and
  Pattern Recognition}, pages 5084--5093, 2020.

\bibitem{pumarola2018unsupervised}
Albert Pumarola, Antonio Agudo, Alberto Sanfeliu, and Francesc Moreno-Noguer.
\newblock Unsupervised person image synthesis in arbitrary poses.
\newblock In {\em Proceedings of the IEEE Conference on Computer Vision and
  Pattern Recognition}, pages 8620--8628, 2018.

\bibitem{qi2021face}
Xingqun Qi, Muyi Sun, Weining Wang, Xiaoxiao Dong, Qi Li, and Caifeng Shan.
\newblock Face sketch synthesis via semantic-driven generative adversarial
  network.
\newblock In {\em 2021 IEEE International Joint Conference on Biometrics
  (IJCB)}, pages 1--8. IEEE, 2021.

\bibitem{ren2020dynamic}
Min Ren, Yunlong Wang, Zhenan Sun, and Tieniu Tan.
\newblock Dynamic graph representation for occlusion handling in biometrics.
\newblock In {\em Proceedings of the AAAI Conference on Artificial
  Intelligence}, volume~34, pages 11940--11947, 2020.

\bibitem{ren2020deep}
Yurui Ren, Xiaoming Yu, Junming Chen, Thomas~H Li, and Ge Li.
\newblock Deep image spatial transformation for person image generation.
\newblock In {\em Proceedings of the IEEE/CVF Conference on Computer Vision and
  Pattern Recognition}, pages 7690--7699, 2020.

\bibitem{ronneberger2015u}
Olaf Ronneberger, Philipp Fischer, and Thomas Brox.
\newblock U-net: Convolutional networks for biomedical image segmentation.
\newblock In {\em International Conference on Medical image computing and
  computer-assisted intervention}, pages 234--241. Springer, 2015.

\bibitem{salimans2016improved}
Tim Salimans, Ian Goodfellow, Wojciech Zaremba, Vicki Cheung, Alec Radford, and
  Xi Chen.
\newblock Improved techniques for training gans.
\newblock {\em Advances in neural information processing systems},
  29:2234--2242, 2016.

\bibitem{shen2018person}
Yantao Shen, Hongsheng Li, Shuai Yi, Dapeng Chen, and Xiaogang Wang.
\newblock Person re-identification with deep similarity-guided graph neural
  network.
\newblock In {\em Proceedings of the European conference on computer vision
  (ECCV)}, pages 486--504, 2018.

\bibitem{siarohin2018deformable}
Aliaksandr Siarohin, Enver Sangineto, St{\'e}phane Lathuiliere, and Nicu Sebe.
\newblock Deformable gans for pose-based human image generation.
\newblock In {\em Proceedings of the IEEE Conference on Computer Vision and
  Pattern Recognition}, pages 3408--3416, 2018.

\bibitem{song2019unsupervised}
Sijie Song, Wei Zhang, Jiaying Liu, and Tao Mei.
\newblock Unsupervised person image generation with semantic parsing
  transformation.
\newblock In {\em Proceedings of the IEEE/CVF Conference on Computer Vision and
  Pattern Recognition}, pages 2357--2366, 2019.

\bibitem{tang2020xinggan}
Hao Tang, Song Bai, Li Zhang, Philip~HS Torr, and Nicu Sebe.
\newblock Xinggan for person image generation.
\newblock In {\em European Conference on Computer Vision}, pages 717--734.
  Springer, 2020.

\bibitem{wang2004image}
Zhou Wang, Alan~C Bovik, Hamid~R Sheikh, and Eero~P Simoncelli.
\newblock Image quality assessment: from error visibility to structural
  similarity.
\newblock {\em IEEE transactions on image processing}, 13(4):600--612, 2004.

\bibitem{wu2020adaptive}
Yiming Wu, Omar El~Farouk Bourahla, Xi Li, Fei Wu, Qi Tian, and Xue Zhou.
\newblock Adaptive graph representation learning for video person
  re-identification.
\newblock {\em IEEE Transactions on Image Processing}, 29:8821--8830, 2020.

\bibitem{yan2018spatial}
Sijie Yan, Yuanjun Xiong, and Dahua Lin.
\newblock Spatial temporal graph convolutional networks for skeleton-based
  action recognition.
\newblock In {\em Thirty-second AAAI conference on artificial intelligence},
  2018.

\bibitem{yan2019learning}
Yichao Yan, Qiang Zhang, Bingbing Ni, Wendong Zhang, Minghao Xu, and Xiaokang
  Yang.
\newblock Learning context graph for person search.
\newblock In {\em Proceedings of the IEEE/CVF Conference on Computer Vision and
  Pattern Recognition}, pages 2158--2167, 2019.

\bibitem{zhang2021pise}
Jinsong Zhang, Kun Li, Yu-Kun Lai, and Jingyu Yang.
\newblock Pise: Person image synthesis and editing with decoupled gan.
\newblock In {\em Proceedings of the IEEE/CVF Conference on Computer Vision and
  Pattern Recognition}, pages 7982--7990, 2021.

\bibitem{zhang2020cross}
Pan Zhang, Bo Zhang, Dong Chen, Lu Yuan, and Fang Wen.
\newblock Cross-domain correspondence learning for exemplar-based image
  translation.
\newblock In {\em Proceedings of the IEEE/CVF Conference on Computer Vision and
  Pattern Recognition}, pages 5143--5153, 2020.

\bibitem{zhang2018unreasonable}
Richard Zhang, Phillip Isola, Alexei~A Efros, Eli Shechtman, and Oliver Wang.
\newblock The unreasonable effectiveness of deep features as a perceptual
  metric.
\newblock In {\em Proceedings of the IEEE conference on computer vision and
  pattern recognition}, pages 586--595, 2018.

\bibitem{zhang2021keypoint}
Shaobo Zhang, Wanqing Zhao, Ziyu Guan, Xianlin Peng, and Jinye Peng.
\newblock Keypoint-graph-driven learning framework for object pose estimation.
\newblock In {\em Proceedings of the IEEE/CVF Conference on Computer Vision and
  Pattern Recognition}, pages 1065--1073, 2021.

\bibitem{zhu2020sean}
Peihao Zhu, Rameen Abdal, Yipeng Qin, and Peter Wonka.
\newblock Sean: Image synthesis with semantic region-adaptive normalization.
\newblock In {\em Proceedings of the IEEE/CVF Conference on Computer Vision and
  Pattern Recognition}, pages 5104--5113, 2020.

\bibitem{zhu2019progressive}
Zhen Zhu, Tengteng Huang, Baoguang Shi, Miao Yu, Bofei Wang, and Xiang Bai.
\newblock Progressive pose attention transfer for person image generation.
\newblock In {\em Proceedings of the IEEE/CVF Conference on Computer Vision and
  Pattern Recognition}, pages 2347--2356, 2019.

\end{thebibliography}
}

\end{document}